\documentclass[acmlarge]{acmart}
\makeatletter
\newcommand{\confshort}{\acmConference@shortname}
\newcommand{\conffull}{\acmConference@name}
\newcommand{\confdate}{\acmConference@date}
\newcommand{\confloc}{\acmConference@venue}
\AtBeginDocument{
  \fancypagestyle{firstpagestyle}{
    \fancyhead{}%
    \fancyfoot[C]{}%
  }
  \fancyhf{}
  \fancyhead[LO]{\@headfootfont\shorttitle}%
  \fancyhead[RE]{\@headfootfont\@shortauthors}%
  \fancyhead[LE]{\@headfootfont\footnotesize \confshort, \confdate, \confloc}%
  \fancyhead[RO]{\@headfootfont\footnotesize \confshort, \confdate, \confloc}%
  \fancyfoot[C]{}%
}
\makeatother
\acmBooktitle{\conffull\@ (\confshort), \confdate, \confloc}
\AtBeginDocument{%
  }

\copyrightyear{2026}
\acmYear{2026}
\setcopyright{cc}
\setcctype{by}
\acmConference[FAccT '26]{The 2026 ACM Conference on Fairness, Accountability, and Transparency}{June 25--28, 2026}{Montreal, QC, Canada}
\acmBooktitle{The 2026 ACM Conference on Fairness, Accountability, and Transparency (FAccT '26), June 25--28, 2026, Montreal, QC, Canada}
\acmDOI{10.1145/3805689.3812287}
\acmISBN{979-8-4007-2596-8/2026/06}

\usepackage[dvipsnames]{xcolor}
\usepackage[table]{xcolor}
\definecolor{lightgray}{gray}{0.92}
\definecolor{lightgreen}{RGB}{220, 240, 220}

\usepackage[table]{xcolor}
\usepackage{tikz} 

\definecolor{mygreen}{RGB}{64,176,166} 
\definecolor{myred}{RGB}{230, 97, 0} 

\newcommand{\female}[1]{\textcolor{RedOrange}{\textbf{#1}}}
\newcommand{\male}[1]{\textcolor{PineGreen}{\textbf{#1}}}

\newcommand{\ShadeCell}[7]{%
  \begingroup
  \pgfmathsetmacro{\val}{#2}%
  \pgfmathsetmacro{\mn}{#3}%
  \pgfmathsetmacro{\mx}{#4}%
  \pgfmathsetmacro{\lo}{#5}%
  \pgfmathsetmacro{\hi}{#6}%
  \pgfmathsetmacro{\raw}{(\mx-\mn)==0 ? \hi : (\lo + (\hi-\lo)*(\val-\mn)/(\mx-\mn))}%
  \pgfmathsetmacro{\clamped}{max(\lo, min(\hi, \raw))}%
  \pgfmathtruncatemacro{\pctint}{\clamped}%
  \edef\XColorPct{\pctint}%
  \expandafter\cellcolor\expandafter{#1!\XColorPct}#7%
  \endgroup
}

\usepackage{xcolor}
\usepackage{soul}

\begin{document}

\title{Neutrality Bites: Gender Representation in AI-Generated Animal Stories}

\author{Imani Finkley}
\email{ifinkley@uw.edu}
\orcid{0009-0005-2323-4461}
\affiliation{%
  \institution{University of Washington}
  \city{Seattle}
  \state{Washington}
  \country{USA}
}

\author{Yuanxi Li}
\email{yuanxili@uw.edu}
\orcid{0009-0007-7034-2726}
\affiliation{%
 \institution{University of Washington}
 \city{Seattle}
 \state{Washington}
 \country{USA}}

\author{Melanie Walsh}
\email{melwalsh@uw.edu}
\orcid{0000-0003-4558-3310}
\affiliation{%
  \institution{University of Washington}
  \city{Seattle}
  \state{Washington}
  \country{USA}}

\

\begin{abstract}
Gender bias in AI-generated stories is a well-documented problem. While much attention has been paid to reducing or mitigating this bias, it is not always clear whether interventions produce genuinely fairer results. To investigate this issue, we examine how large language models (LLMs) handle gender assignment in a narrative context that is popular, highly ambiguous, and also known to closely reproduce human stereotypes: stories about talking animals. We prompt six leading LLMs to complete an English-language story about seven different anthropomorphic animal characters whose gender is unstated. We additionally iterate with four different narrative settings and a range of model temperatures. Across the 23.8K stories, we find that models frequently avoid gendering the animal character in the story (19\% on average) or use gender-neutral language like “it” or “its” (38.2\% on average). However, when gender is assigned, there is a significant masculine bias. Feminine animal characters are virtually absent, present in just 2.2\% of stories vs. 40.6\% that feature masculine characters. Our findings point to a broader argument: neutrality bites. In other words, models that prioritize
neutrality to address social bias may actually contribute to the erasure of marginalized perspectives and identities. We suggest that alternative strategies beyond neutrality need to be pursued, such as ones that more equally distribute social possibilities across imagined subjects.
\end{abstract}




\maketitle
\begin{figure}
    \centering
    \includegraphics[width=0.9\linewidth]{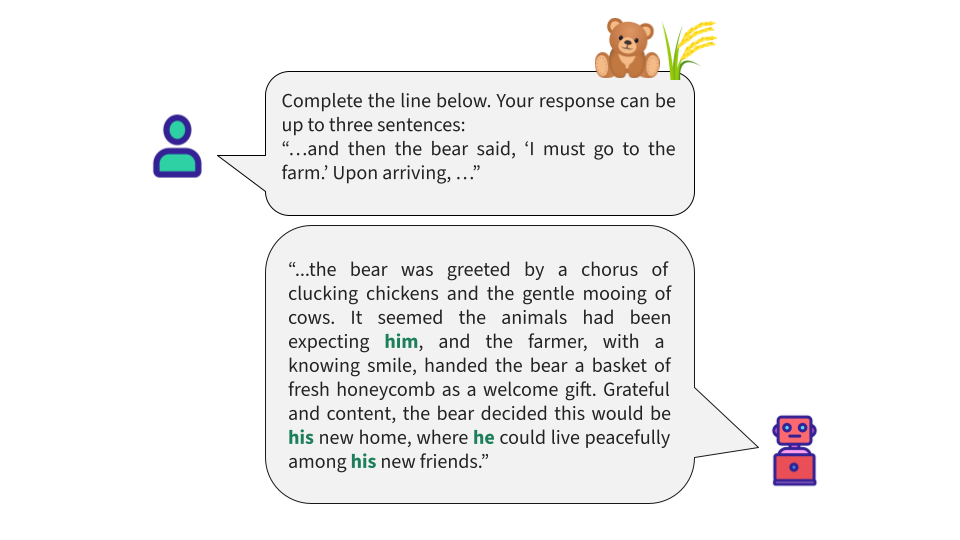}
    \caption{Example prompt and response from GPT-4o with the following parameters: bear (subject), farm (setting), and 1.0 (temperature). We label this story \textcolor{PineGreen}{Masculine} due to the pronouns (bold green) used to refer to the bear. Generated December 2025.}
    \label{fig:story}
\end{figure}
\section{Introduction}
Gender bias in AI-generated outputs has been a widespread problem \cite{sheng-etal-2019-woman, 10.1145/3442188.3445922, lucyGenderRepresentationBias2021}. These systems frequently overrepresent cisgender men and masculine perspectives while rendering women, non-binary people, and femininity either less visible, or reductively and offensively portrayed \cite{10.1145/3582269.3615599, blodgettLanguageTechnologyPower2020}. 
This is concerning not only for algorithmic decision-making contexts like hiring, but also for storytelling and fiction, where AI models are increasingly used. For example, Google's Gemini ``Storybook'' tool \cite{googleGeminiStorybookStories}, which creates 10-page illustrated narratives based on prompts, is now being used by parents to read stories to their children \cite{osullivanGoogleLaunchesPersonalized}. 

Reducing or mitigating gender bias in large language models (LLMs) is now an active area of research in both academia and industry  \cite{zhang-etal-2025-genderalign, ghanbarzadeh-etal-2023-gender, watson-etal-2025-analyzing}. 
But it is not always clear whether proposed or adopted interventions produce genuinely fairer representations, especially for fiction. 
For instance, if a user prompts an LLM to generate a story about a character whose gender is ambiguous, and the model avoids gendering the subject, is that real fairness? Or is it a band-aid fix that covers up a larger bias problem, leaving it to seep out elsewhere? 
To investigate this issue, we examine how LLMs handle gender assignment in a narrative context that is popular, highly ambiguous, and also known to closely reproduce human stereotypes: stories about talking animals (Figure \ref{fig:story}).
We turn to anthropomorphic fiction for several reasons.

First, animal stories are popular and influential.
Human authors have been writing stories about anthropomorphic animals for thousands of years, back to \textit{Aesop's Fables} (620-564 BCE) and beyond.
Today, they feature prominently in children's books and media—like \textit{Pete the Cat} (2008-present) or \textit{Bluey} (2018-present)—which are important for human development and shape early conceptions of gender and gender roles.
Additionally, in the last several years, users have turned to AI models to create children's books  \cite{Lee_2022, 10.1145/3711006, 10.1145/3290607.3312965}, such as \textit{Alice and Sparkle}, one of the first known children’s books to include AI-generated text and images \cite{reshiAliceSparkle2022}.

Second, anthropomorphic narratives represent a unique projection and reflection of human culture.
Animal characters that resemble humans offer useful imaginative affordances for creators.
Some authors reportedly turn to animal characters to conjure ``universal'' subjects who ``transcend'' gender, race, or other identity categories and physical characteristics \cite{mccabeGenderTwentiethCenturyChildrens2011, yabroffWhyAreThere2016}.
Yet, counterintuitively, research shows that gender bias is actually more pronounced in stories about animal characters than in stories about human characters \cite{caseySixtyYearsGender2021, mccabeGenderTwentiethCenturyChildrens2011}.
In other words, paradoxically, human writers project human stereotypes more strongly in animal stories, making them a striking test case for LLMs.

Third, we hypothesize that model developers have not created guardrails or aligned models in response to this specific context, as they almost certainly have for stories about humans in different occupations, like doctors or flight attendants, a genre that has received more scholarly and popular attention \cite{10.1145/3582269.3615599, dimgbaMitigationGenderEthnicity2025, lucyGenderRepresentationBias2021, spillnerUnexpectedGenderStereotypes}.

In our experiment, we iteratively prompt six leading LLMs to complete an English-language story about seven different human-like animal characters whose gender is unstated---bear, dog, rabbit, cat, mouse, pig, and bird.
We additionally iterate with four different narrative settings and a range of model temperatures, which are known to influence the randomness or creativity of the output.
For each story, we identify the pronouns used to describe the animal character, which we take as a proxy for gender assignment.

Across the 23.8K LLM-generated stories, we seek to answer the following questions:
\begin{itemize}
    \item \textbf{RQ1:} How often do LLMs assign \male{masculine} (he/him), \female{feminine} (she/her) \textbf{neutral} (it/its, they/them), or \textbf{no pronouns} (animal name) to the fictional animal character, and how does this compare across models? 
    \item \textbf{RQ2:} How do different factors—model, model temperature, animal character, setting—impact gender assignment, if at all?
    \item  \textbf{RQ3:} How do LLMs' pronoun assignments compare to human responses from prior work?
\end{itemize}

Overall, we find that models frequently avoid gendering the animal character in the story (\textbf{19\%} on average) or use gender-neutral language like ``it'' or ``its'' (\textbf{38.2\%} on average). 
However, when gender \textit{is} assigned, there is a significant masculine bias.
Feminine animal characters are virtually absent, present in just \textcolor{RedOrange}{\textbf{2.2\%}} of stories vs.  \textcolor{PineGreen}{\textbf{40.6\%}} featuring masculine characters.
They mostly appear in stories about cats (\textcolor{RedOrange}{\textbf{53.2\%}}), which are stereotypically associated with femininity. 
Even the model with the highest feminine representation, Claude Sonnet 4.5, only features feminine characters in \textcolor{RedOrange}{\textbf{3.8\%}} of the stories vs. \textcolor{PineGreen}{\textbf{34.3\%}} masculine stories.
Compared to a human study that used an almost identical story prompt \cite{walsh_bearswillbeboys}, LLMs were \textcolor{RedOrange}{\textbf{six times \textit{less} likely}} to imagine a feminine character and \textbf{1.2 times more likely} to use neutral or animal name references.



Notably, gender assignment is not consistent across LLMs, including those from the same developer.
Surprisingly, the most recent OpenAI model, GPT-5.1, displayed the strongest masculine bias, with \textcolor{PineGreen}{\textbf{65.2\%}} of the stories featuring masculine characters.
Though differences in pretraining data or architecture cannot be ruled out, these observed disparities suggest that post-training interventions likely play a role in shaping gendered behavior in these models.
Model temperature and narrative setting did not have major impacts on gender assignment.

Our findings point to an overarching claim: \textbf{\textit{neutrality bites}}. 
We argue, in other words, that models that prioritize neutrality in order to address social bias may unintentionally contribute to 
the erasure of marginalized perspectives and identities.
We suggest that alternative strategies beyond neutrality need to be pursued, such as ones that more equally distribute gender possibilities across imagined subjects.  
Code and materials to support this work can be found at 
\href{https://github.com/imanif/animal-stories}{https://github.com/imanif/animal-stories}.




\section{Related Work}



\subsection{Gender Bias in Children's Literature}
Anthropomorphic animals, like a bear who lives in a house or a tiger that talks, feature prominently in children's literature, a widely consumed genre that plays a formative role in establishing social norms and frameworks for kids \cite{berryGenderedPortrayalInanimate2017}.
A large body of research has pointed to significant gender bias in English-language children's literature, such as the overrepresentation of masculine characters, especially as protagonists \cite{lewisWhatMightBooks} or in active storylines \cite{adukiaWhatWeTeach2023}, and the reinforcement of a gender binary \cite{nevesCanGenderNouns2023, hillMouseLooksBoy2020, careFemaleAnimalCharacters2024, grauerholzGenderRepresentationChildrens1989}.
While \citet{mccabeGenderTwentiethCenturyChildrens2011} found that the representation of feminine human characters steadily improved throughout the 20th century, they found that representation of feminine animal characters did not, remaining stagnant. 
\citet{walsh_bearswillbeboys} found that gender bias persisted among a selection of the most popular and widely-read English-language children's books featuring anthropomorphic animals, even into the 21st century. This concern also extends beyond books themselves to the broader information environments; for example, children's products are often retrieved and suggested through stereotypically gendered associations \cite{rajFireDragonUnicorn2022}.

\subsection{Gender Bias in LLM-Generated Fiction}
Gender bias is a well-demonstrated and widely discussed problem in LLMs \cite{sheng-etal-2019-woman, 10.1145/3442188.3445922, hossain-etal-2023-misgendered}. Notably, narrative texts---especially short stories---are frequently used as test cases for audits of gender and social bias, even in research not primarily concerned with fiction or narrative form \cite{gillespieGenerativeAIPolitics2024, lucyGenderRepresentationBias2021, shiehLaissezFaireHarmsAlgorithmic2025}. 
As with human-authored literature, stories serve as a useful window into broader cultural imaginaries: they make visible assumptions, norms, and stereotypes, and they can expose implicit biases that may remain less visible in more constrained or task-oriented settings, like the review of job applications. 

Many studies that explicitly investigate LLM-generated stories or use them as test cases have found significant gender bias. 
For example, \citet{lucyGenderRepresentationBias2021} found thematic biases in GPT-3-generated stories, showing that feminine characters were primarily described by their appearance while masculine characters were
associated with high power.
\citet{spillnerUnexpectedGenderStereotypes} similarly found that, in ChatGPT stories, women were overrepresented in stereotypically women-dominated fields---though also found, in a reversal, that women were overrepresented in some stereotypically \textit{male}-dominated fields, perhaps reflecting alignment efforts to avoid bias.
A study that used LLMs to produce children's bedtime stories also found stereotypical gender representations, as well as limited diversity in the socioeconomic backgrounds and ethnicity of characters in narratives given different geographical settings  \cite{rooeinBiasedTalesCultural2025a}. 

Perhaps most closely related to our own work, \citet{gillespieGenerativeAIPolitics2024} found that leading models from Microsoft and OpenAI often ``avoided'' gendering the subject in an ambiguous story, speculating that ``this is almost certainly the result of interventions meant to address the uneasy cultural politics around pronouns and gender identity.''


\subsection{Gender Debiasing \& Finetuning}
Attempting to ``debias'' LLMs has been a growing trend in response to this evidence, and it has taken a variety of forms, in recognition of the array of harms to various gender minority groups \cite{caoGenderInclusiveCoreferenceResolution2021, hossain-etal-2023-misgendered}. Proposed interventions include finetuning models to correctly employ neopronouns---such as, in English, xe/xem/xyr or ze/zir/zirs, or, in Dutch, hen/hen/hun or die/die/diens---to reflect a category of pronoun used by some non-binary people \cite{vanbovenTransformingDutchDebiasing2024}.
There is also work on so-called ``gendertuning,'' which uses bias word perturbation and fine-tuning classification to revise texts with non-gendered pronouns or pronouns that disrupt common stereotypes (e.g. ``women in the kitchen'' becomes ``men in the kitchen'') \cite{ghanbarzadeh-etal-2023-gender}. Adopting generated-language reforms remains a challenge. 
For example, when tasked with explaining their text revisions, LLMs have been found to treat gender-inclusivity as more relevant for women or trans people, rather than as a default approach \cite{watson-etal-2025-analyzing}. 

Recent work further suggests that apparent gender neutrality does not necessarily translate into equitable representation. \citet{mickel2026more} show that increased representation of women in model outputs does not eliminate differences in how women and men are described, while \citet{10.1007/978-3-031-56069-9_1} show that seemingly balanced results can still encode biased exposure patterns. Together, these studies suggest that fairness must be evaluated not only in terms of who appears in outputs, but also in terms of how gendered subjects are described and distributed. A broader perspective from information retrieval is also useful here. \citet{seyedsalehiMitigatingGenderBias2025} show that gender bias can emerge at multiple stages of an information system, including queries, datasets, ranking, and evaluation, which helps frame gender neutrality not simply as a matter of wording, but as a broader systems problem.


\section{Methods}
\subsection{Story Generation}

To create our animal story dataset, we prompt six diverse, state-of-the-art LLMs 
to complete a specific English-language story template.
We test two generations of OpenAI models---GPT-4o \cite{openai_gpt4_system_card} and GPT-5.1 \cite{openai_gpt5_system_card}---and leading models from Anthropic and Google---Claude Sonnet 4.5 \cite{anthropic_claude4_system_card} and  Gemini 2.5 Flash \cite{comanici2025gemini}---as well as two widely used open or hybrid models---Mistral Medium 3.1 \cite{MistralMedium31} and OLMo3 7B Instruct \cite{olmo2025olmo3}.
We prompt each model with an in-media-res story, introducing a talking animal character who speaks in the first person and is not clearly gendered in any way:
\begin{quote}
  \textit{  Complete the line below. Your response can be up to three sentences: 
    \newline
    ``…and then the \textbf{\{animal\}} said, `I must go to the \textbf{\{setting\}}.' Upon arriving, …''}
\end{quote}

We adopt this template from a previous study conducted by the senior author with journalists from \textit{The Pudding}, in which they gave the same prompt to human survey participants \cite{walsh_bearswillbeboys}. This allows us to compare model behavior to human responses from prior work. We systematically vary the story with seven different animals and four different settings, in order to account for biases that may be introduced with any particular animal or location. The animal characters were selected based on the same prior work \cite{walsh_bearswillbeboys}, which included a survey of 300 popular English-language children's books on \textit{Goodreads} that feature at least one anthropomorphized animal, focusing on the most rated titles published since 1950, plus some older and still highly rated classics. We also noted settings commonly used in these narratives and selected four settings with potentially embedded cultural stereotypes (e.g., domestic settings like a kitchen may be more associated with feminine characters, while outdoor spaces like a farm may be more associated with masculine characters) for our experiment design. 
Additionally, we vary the temperature for each model, experimenting with the model's default temperature as well as one temperature below and above it.\footnote{The default temperature for Claude Sonnet 4.5 is also the maximum temperature---1.0---making it impossible to test at a temperature above the default. Thus, for the Claude model, we only prompt at the default temperature and one temperature below it.}
We prompt each model 50 times with each animal, setting, and temperature combination, totaling 23.8K stories and approximately 1.05M tokens (Table \ref{tab:total}).

\begin{table}[h]
  \centering
  \begin{tabular}{l l}
  \toprule
  Story Variable &  \\
  \midrule
  Animals (7) & bear, bird, cat, dog, mouse, pig, rabbit \\
  Settings (4) & farm, kitchen, river, store \\
  \bottomrule
  \end{tabular}
  \caption{Experimental design: Animals and settings used for story generation prompts.}
  \label{tab:design}
  \end{table}



  \begin{table}[h]
  \centering
  \label{tab:stories}
  \begin{tabular}{lrl}
  \toprule
  Model & Stories & Temperatures \\
  \midrule
  Claude Sonnet 4.5 & 2,800 & 0.75, \textbf{1.0} \\
  Gemini-2.5 & 4,200 & 0.75, \textbf{1.0}, 1.25 \\
  GPT-4o & 4,200 & 0.75, \textbf{1.0}, 1.25 \\
  GPT-5.1 & 4,200 & 0.75, \textbf{1.0}, 1.25 \\
  Mistral Medium & 4,200 & 0.5, \textbf{0.7}, 0.9 \\
  OLMo3 & 4,200 & 0.75, \textbf{1.0}, 1.25 \\
  \midrule
  \textbf{Total} & \textbf{23,800} & \\
  \end{tabular}
  \caption{Stories generated per model. 1.4K stories generated per temperature. Bold indicates default temperature for each model.}
   \label{tab:total}
  \end{table}

\subsection{Story Analysis}
To determine the gender of the animal in the story, we rely on pronouns in the text.
In the English language, pronouns often imply the gender of a subject.
We establish the following categories: \textbf{\textcolor{RedOrange}{Feminine}} (she, her, hers, herself); \textbf{\textcolor{PineGreen}{Masculine }} (he, him, his, himself); \textbf{Neutral} (it, its, itself; they, them, their, theirs, themself, themselves). 
Cases in which no pronouns are used, and the character is simply referred to by the animal's name (e.g., ``the pig'') are labeled \textbf{Animal Name}. We recognize that this is not an exhaustive classification of gender but present these labels in an effort towards examining gender representation in generated literature and NLP that extends beyond a binary \cite{caoGenderInclusiveCoreferenceResolution2021, devinneyTheoriesGenderNLP2022}.

In some of our analyses, we combine \textbf{Neutral} and \textbf{Animal Name} into an overarching gender-neutral assignment category to highlight the impact of not assigning a character a specific gender. 
However, we do not treat these outcomes as conceptually identical.
Especially in human contexts, \textbf{neutral} pronouns like ``they/them'' are often an intentional linguistic choice aligned with inclusivity or with recognition of non-binary gender expressions.
We acknowledge that \textbf{neutral} language, and even gender-ambiguous language, is sometimes desirable and carries positive, socially progressive connotations. 
Two considerations shape how we handle this complexity. 
First, the models rarely used ``they/them'' pronouns for animal characters and instead used ``it/its'' in a manner that we view as similar to the use of the animal's name.
Consider, for example, a model completion where a pig is referred to as ``it'' versus ``the pig'': 



\begin{quote}
    ..and then the pig said, `I must go to the river.' Upon arriving, \textbf{it} gazed at the rippling waters and sighed contentedly (OLMo3, $animal$ = pig, $temperature$ = 1.25, $setting$ = river)
\end{quote}


\begin{quote}
    ...and then the pig said, `I must go to the river.' Upon arriving, \textbf{the pig }saw a group of frogs gathered on lily pads, waiting for rain to arrive. (OLMo3, $animal$ = pig, $temperature$ = 1.25, $setting$ = river)
\end{quote}
We read these two stories as functionally similar in their representation of the pig's gender, or lack thereof, though we note that the use of ``it'' is perhaps more intentionally non-gendered and in this context may objectify or even de-humanize the pig more.
This leads to our second point. Because we believe this difference is small but still meaningful to preserve, we adopt a dual approach. We report results with each specific category noted, and we also present our analysis with a generalized neutrality label to investigate broader patterns in model behavior.



To parse which pronouns refer to the animal character specified in the prompt, we use coreference resolution. 
Coreference resolution disambiguates references to subjects using contextual evidence.
This method is necessary to analyze the stories because ambiguity can be introduced with multiple characters or objects---e.g. ``The \textbf{bear} and the \textbf{farmer} were in the store and saw a \textbf{jar of honey}. The \textbf{farmer} asked \female{her} if \female{she} wanted \textbf{it}.''
We need to identify the pronouns related to the main animal character---e.g. the bear (``she/her'')---not other characters or objects---e.g. the jar of honey (``it'').
To identify these references, we specifically use the FastCoref package \cite{otmazgin-etal-2022-f}.
We manually annotated 200 randomly sampled stories and measured how our labels compared to FastCoref. Across each gender category, FastCoref achieved a mean F1 score of 0.95 (see more details in Appendix \ref{coref}). We then manually reviewed stories labeled Animal Name or Neutral to minimize false positives given the ambiguous nature of this language.



\subsection{Establishing a Baseline}

\begin{figure}[h]
    \centering
    \includegraphics[width=0.75\linewidth]{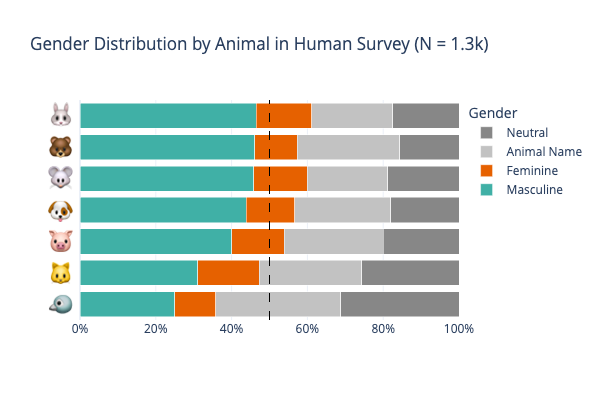}
    \caption{Gender distribution results from 1327 human survey respondents given a story-completion task with varying anthropomorphized-animal protagonists.}
    \Description{Bar chart}
    \label{fig:survey}
\end{figure}

To establish a human baseline for the LLM-generated stories, we draw on prior work that surveyed approximately 1.3K respondents (see details in Appendix \ref{survey-dems}). Respondents were asked to complete a nearly identical story to the one provided to the LLMs \cite{walsh_bearswillbeboys}:\footnote{The survey was circulated to readers of \textit{The Pudding}, a data journalism outlet, on social media and via email, offering an opportunity for respondents to win free merchandise.}

\begin{quote}
   \textit{ In the 1980s, there was a study that produced surprising results about our predictive tendencies. We’re conducting a similar study today.}
    
   \textit{ Instructions: Complete the line below. What did the \textbf{\{animal\}} do? Your response can be up to three sentences.}
    
  \textit{ ``...and then the \textbf{\{animal\}} said, `I must go to the river.' Upon arriving,...''}
\end{quote} 
The initial part of the prompt—misdirecting the respondent about the focus of the study—was included to avoid priming participants to explicitly consider gender. 
Survey respondents were randomly given one of seven animals: bear (204 stories), bird (176 stories), cat (203 stories), dog (189 stories), mouse (192 stories), pig (180 stories), or rabbit (183 stories).
Of the 1,327 completed stories, \textcolor{PineGreen}{\textbf{39.9 percent}} assigned masculine pronouns to animal protagonists, while \textcolor{RedOrange}{\textbf{13.4 percent}} of the stories employed feminine pronouns. 
Cat had the most feminine assignments (\textcolor{RedOrange}{\textbf{16.3 \%}}), while bird had the most neutral representation ({\textbf{64.3\%}}) (see Figure \ref{fig:survey}).

To compare the model-generated narratives with our human baseline, we restrict our analysis to a subset of the LLM-generated stories where the $setting$ is river (\textit{n} = 5950). Since the dependent variable, \textit{Gender} \{Masculine, Feminine, Neutral, Animal Name\}, is a nominal categorical variable with four levels, we use a mixed multinomial logistic regression model, which allows us to estimate differences across gender categories and accounts for multiple outputs from each model. 

We treat this as a 2x7 mixed factorial design with the following fixed factors: \textit{Author} \{LLM, Human\} and \textit{Animal} \{bear, bird, cat, dog, mouse, pig, rabbit\}, allowing us to examine both their effects and interactions on gender assignment. We then used an analysis of variance, implemented with the multinomial Poisson transformation \cite{Baker1994}, to identify any statistically significant differences between the survey respondents and the LLMs (averaged across temperature and model variations to reduce within-model variability).

\section{Results}
\subsection{Overall}


Across the 23.8K LLM-generated stories, we find, overall, that the six models predominantly either avoided gendering the anthropomorphic animal character or gendered the character with neutral pronouns like ``it'' (\textbf{57.2\% total}). The language used for gender-neutral representation did not vary greatly: only \textbf{two} stories used ``they/them/theirs'' to refer to a single animal character, once for a human side character.
However, when the animal protagonist was gendered, it was gendered masculine \textcolor{PineGreen}{\textbf{95 percent}} of the time (Table \ref{tab:model-table}). Only \textcolor{RedOrange}{\textbf{2.2\%}} of the outputs featured a feminine animal character---that is, just \female{513} of \textbf{23,800 stories} (compared to \male{9,673} masculine stories).

\begin{table}[h]
    \centering
    \begin{tabular}{lccc}
        \hline
        Model & Neutral or Animal Name \% & Masculine \% & Feminine \% \\
        \hline
        OLMo3             
            & \cellcolor{gray!34}\textbf{85.3}
            & \cellcolor{mygreen!5}11.5
            & \cellcolor{myred!6}3.2 \\
        Mistral Medium     
            & \cellcolor{gray!29}\textbf{71.9}
            & \cellcolor{mygreen!11}27.9
            & \cellcolor{myred!1}0.2 \\
        Claude Sonnet 4.5 
            & \cellcolor{gray!25}\textbf{62.0}
            & \cellcolor{mygreen!14}34.3
            & \cellcolor{myred!7}3.8 \\
        GPT-4o           
            & \cellcolor{gray!23}\textbf{57.0}
            & \cellcolor{mygreen!17}40.2
            & \cellcolor{myred!6}2.8 \\
        Gemini-2.5        
            & \cellcolor{gray!15}35.6
            & \cellcolor{mygreen!22}\textbf{62.7}
            & \cellcolor{myred!4}1.7 \\
        GPT-5.1         
            & \cellcolor{gray!13}33.0
            & \cellcolor{mygreen!26}\textbf{65.2}
            & \cellcolor{myred!4}1.8 \\

        \hline
         \textbf{Human Survey Baseline}
            & \cellcolor{gray!19}\textbf{46.7}
            & \cellcolor{mygreen!16}\textbf{39.8}
            & \cellcolor{myred!18}\textbf{13.4} \\
        \textbf{Model Average}
            & \cellcolor{gray!22}\textbf{57.2}
            & \cellcolor{mygreen!18}\textbf{40.6}
            & \cellcolor{myred!4}\textbf{2.2} \\

        \hline
    \end{tabular}
    \caption{Gendered Pronoun Breakdown in Animal Stories Across LLMs. Human survey baseline drawn from a survey by \textit{The Pudding}.}
    \label{tab:model-table}
\end{table}

\begin{figure}[h!]
    \includegraphics[width=0.85\linewidth]{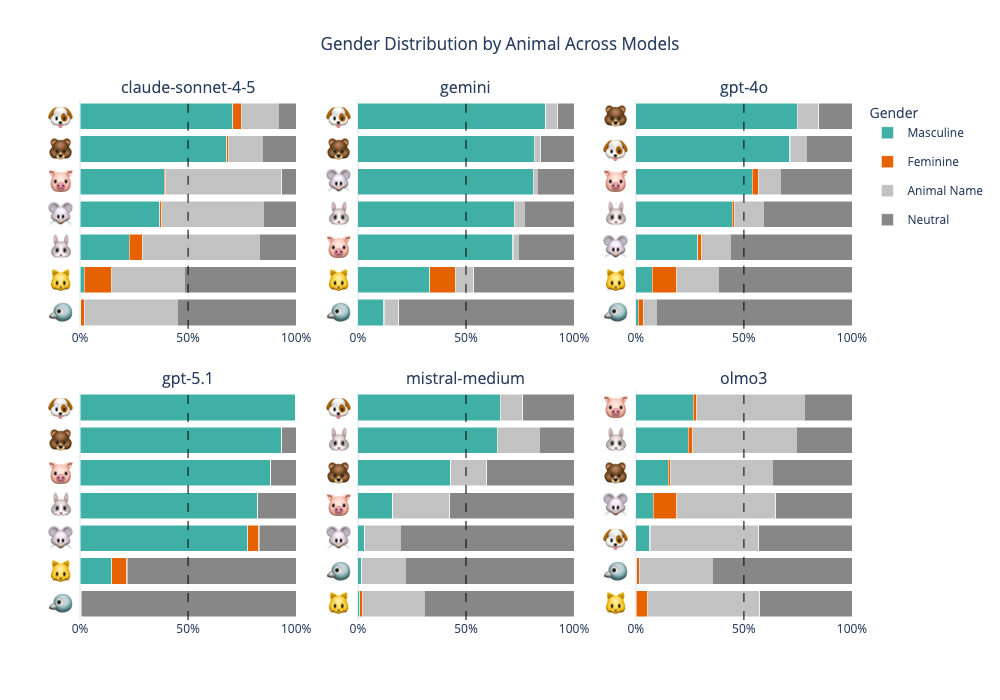}
    \caption{Gender assignment distribution results from 23.8K text outputs from six LLMs given a story-completion task with varying anthropomorphized-animal protagonists, narrative settings, and temperature parameters.}
    \label{fig:models-animals}
\end{figure}

The most ``neutral'' model, OLMo3, also had the second-best feminine representation.
OLMo repeated the animal's name, rather than assigned a gender via pronouns, in stories \textbf{46.6 percent} of the time, while using neutral pronouns \textbf{38.6 percent} of the time.  
Only \female{3.2 percent}---or \female{135} of \textbf{4,200 stories}--- contained feminine main characters, with \male{11.5 percent} containing masculine main characters.

Mistral rarely cast feminine characters. Only \female{9} stories (\female{0.2 \%}) contained feminine protagonists. Moreover, all of these feminine characters were cats. Conversely, Mistral used gender-neutral pronouns for its characters for \textbf{52.2 percent} of its stories, avoided gender \textbf{19.7 percent} of the time, and employed masculine pronouns for \male{27.9 percent} of the time.

Of all the models, GPT-5.1 and Gemini-2.5 displayed the strongest masculine bias.
The animal protagonist was cast as masculine in \male{65.2 percent} of stories generated by GPT-5.1, far outstripping the masculine proportion of similar stories completed by humans (\male{39.8 \%}).
Outputs were especially skewed depending on the animal.
Stories were more than \male{77 percent} masculine for dogs, bears, pigs, rabbits, and mice.
But cats, stereotypically associated with feminine characters, were flipped.
Cats were not gendered or were described neutrally \textbf{78.3\%} of the time, yet when explicitly gendered, were still predominantly masculine (\male{14.7\%} vs. \female{7\%}). By contrast, birds were\textbf{ 99.3\%} neutral, referenced as ``it'' or ``its.'' 

Despite having a relatively low proportion of feminine characters (\female{3.8\%}), Claude was the only model to produce feminine characters across all seven animal categories, including those typically labeled masculine by human participants. 

\subsection{Compared To The Baseline}
We find a statistically significant difference in how LLMs and the survey respondents resolve gender in ambiguous contexts ($\chi^2$(3, \textit{N}=7277) = 332.63, \textit{p} < 2.2$e^{-16}$). The animal protagonist also had a significant effect on gender assignment distribution in the completed stories ($\chi^2$(18, \textit{N}=7277) = 359.45, \textit{p} < 2.2$e^{-16}$).  Additionally, multinomial logistic regression revealed a significant interaction effect between story author (LLM vs. human) and the talking animal characters ($\chi^2$(18, \textit{N}=7277) = 218.30, \textit{p} < 2.2$e^{-16}$), suggesting that there were distinct differences between how specific protagonists were described versus others. These results indicate that the gender label for a given story depends on both the species and the author. Further, \textit{post hoc} pairwise comparisons, adjusted with Holm's sequential Bonferroni procedure \cite{Holm1979}, revealed significant differences between how humans and the models portrayed cat, especially: LLMs show a stronger neutrality bias for cats (\textbf{88.6 percent}), whereas human respondents produced more balanced gender assignment distributions (\textbf{52.7 percent}) (Table \ref{tab:model-human}). This demonstrates that models encode gender associations across animal protagonists significantly different from observed human behavior.

\begin{table}[h]
    \centering
    \begin{tabular}{lcc}
        \hline
        Animal & LLM Neutral or Animal Name \% & Human Survey Neutral or Animal Name \% \\
        \hline
        Bear        
            & \cellcolor{gray!26}\textbf{54.5}
            & \cellcolor{gray!15}42.6 \\
        Bird             
            & \cellcolor{gray!51}\textbf{98.7}
            & \cellcolor{gray!34}\textbf{64.2} \\
        Cat     
            & \cellcolor{gray!42}\textbf{88.6}
            & \cellcolor{gray!24}\textbf{52.7} \\
        Dog 
            & \cellcolor{gray!10}35.9
            & \cellcolor{gray!12}43.3 \\
        Mouse 
            & \cellcolor{gray!32}\textbf{66.5}
            & \cellcolor{gray!11}40.2 \\
        Pig            
            & \cellcolor{gray!30}\textbf{62.5}
            & \cellcolor{gray!12}46.1 \\
        Rabbit           
            & \cellcolor{gray!24}\textbf{52.7}
            & \cellcolor{gray!10}38.8 \\

        \hline
         \textbf{Total \%}
            & \cellcolor{gray!32}\textbf{65.6}
            & \cellcolor{gray!16}\textbf{46.7} \\

        \hline
    \end{tabular}
    \caption{\textbf{Neutral} and \textbf{Animal Name} distribution across animals in a subset of LLM-generated stories (n=5950) and stories from human survey baseline (N=1327). For all stories, $setting$ = river.}
    \label{tab:model-human}
\end{table}

\subsection{Across Animals}

Across all the models, on average, the top three most neutral animal characters were \textbf{birds} (\textbf{96.3 percent}), \textbf{cats} (\textbf{81.7 percent}), and \textbf{pigs} \textbf{(49.2 percent}), while  
the three most masculine characters were \male{dogs} (\male{66.6 percent}), \male{bears} (\male{62.3 percent}), and \male{rabbits} (\male{53.5 percent}). 
Due to the high presence of neutral characterizations for all the animals, these percentages do not fully communicate how overwhelmingly masculine these animals were imagined.
Out of 3,400 stories each, rabbits had \female{43} feminine portrayals, dogs had \female{19}, while bears had only \female{8}. 
This stands in stark contrast to the human responses, where dogs (\female{12.7\%}), rabbits (\female{14.8\%}), and bears (\female{11.2\%}) were cast as feminine much more often.

\subsection{Across Settings}
We explore the impact of four different narrative settings on the gender assignment in the story.
These settings carry a range of cultural associations and assumptions that may shape the model's narrative choices. 
For example, \textit{kitchen} and \textit{store} are more industrial and potentially human-centered, while \textit{farm} and \textit{river} are more natural and animal-centered. 
From a different angle, \textit{kitchen} and \textit{store} may be stereotypically considered more feminine spaces, because they are more domestic, while \textit{farm} and \textit{river} may be considered more masculine spaces, because they are outdoors and associated with manual labor.

\begin{table}[h]
    \centering
    \begin{tabular}{c c c c c}
        \hline
         Setting & Neutral or Animal Name \% & Masculine \% & Feminine \% \\
         \hline
         River & \cellcolor{gray!27}\textbf{65.6} & \cellcolor{mygreen!14}31.6 & \cellcolor{myred!5}2.8 \\
         Kitchen & \cellcolor{gray!24}\textbf{58.2} & \cellcolor{mygreen!17}40.0 & \cellcolor{myred!1}1.8 \\
         Store & \cellcolor{gray!20}\textbf{52.6} & \cellcolor{mygreen!24}45.8 & \cellcolor{myred!1}1.6 \\
         Farm & \cellcolor{gray!19}\textbf{52.4} & \cellcolor{mygreen!25}45.2 & \cellcolor{myred!5}2.4 \\
          \hline
         \textbf{Model Avg.}
         & \cellcolor{gray!23}\textbf{57.2}
            & \cellcolor{mygreen!16}\textbf{40.6}
            & \cellcolor{myred!5}\textbf{2.2} \\
         \hline
    \end{tabular}
    \caption{Gendered pronoun distribution in animal stories across four narrative settings, averaged across all six LLMs.}
    \label{tab:setting}
\end{table}

Table \ref{tab:setting} shows that gender-neutral assignments were distributed slightly above random across all four narrative settings (\textbf{at least 52 percent}). Farm stories had the most masculine characters (\male{45.2\%}), though not by a large margin.
Most feminine representation was in river narratives (\female{2.8\%}), though, again, not by a significant margin.  

However, this behavior is not consistent across models (see Figure \ref{fig:settings} in Appendix \ref{appendix}). 
For example, in Mistral outputs, \female{77.7 percent} of feminine portrayals are set on a farm, with only \female{two} feminine main characters in a river narrative, \female{zero} in a kitchen narrative, and \female{zero} in a store narrative.

\subsection{Across Temperatures}
The temperature parameter for LLMs controls the degree of randomness for the next generated token, often construed as a creativity parameter. 
The gendered-pronoun distribution for each model did not vary by significant margins when prompted at low, default, or high temperatures  
(see Figure \ref{fig:temps} in Appendix \ref{appendix}).

\section{Discussion}

\begin{figure}[h!]
    \includegraphics[width=0.75\linewidth]{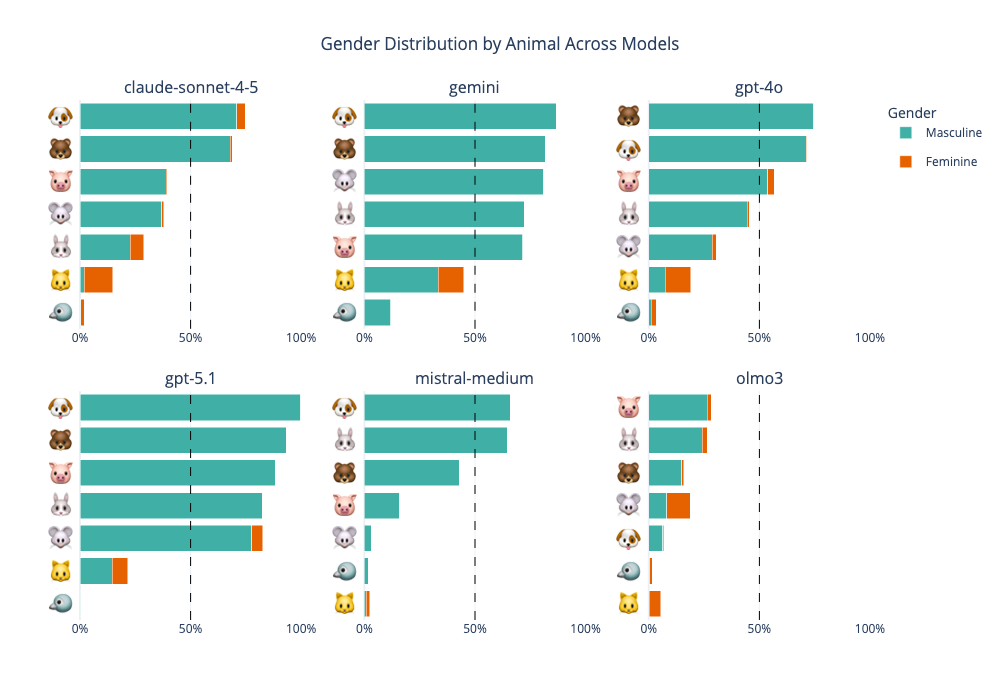}
    \caption{Gender assignment distribution results when gender-neutral representation is not considered, highlighting a significant masculine bias.}
    \label{fig:models-animals-just-mf}
\end{figure}
Our experiment reveals an important pattern in how leading LLMs respond to, or resolve, gender ambiguity in stories featuring anthropomorphized characters.
We show that models will often avoid gendering a subject (``the dog''), or use gender-neutral language (``it/its''), if the subject's gender is ambiguous.
Overall, the LLMs produced more gender-neutral and ambiguously gendered protagonists than the human survey participants (\textbf{57.2\% } vs. \textbf{46.7\%}, respectively).
And the two animals that were most feminine in the human results---birds and cats---were ascribed with gender-neutral pronouns by the  models in the same narrative setting (river) \textbf{86.6} and \textbf{58.1 percent} of the time.
We speculate that gender neutrality may be used by model developers as a guardrail or alignment tactic to counteract known model gender bias.
 
Our findings show that overreliance on neutrality may in fact \textit{exacerbate} gender bias, at least in fictional stories, amplifying the presence of masculine characters and nearly erasing feminine ones. 
Figure \ref{fig:models-animals-just-mf} displays what the masculine vs. feminine breakdown looks like for each animal character with neutrality's ``bite'' taken out of it---in other words, what the distribution looks like if we only consider stories where gender is explicitly assigned, removing the distorting or distracting influence of neutral characterizations.
The picture is bleak.
While the models cast \male{40.6\%} of the characters as masculine, on average, they only cast \female{2.2\%} of the characters as feminine---just \female{513} of 23,800 stories (compared to \male{9,673} masculine stories).
LLMs were \female{six times \textit{less} likely} to create a feminine animal character than human respondents.
And some animals were almost never cast as feminine across any of the models or prompt combinations, like dogs (\female{19} out of 3,400 stories), rabbits (\female{43}  out of 3,400 stories), and bears (\female{8}  out of 3,400 stories).

This erasure matters. 
Stories, even fictional ones about talking animals, have consequences in the world.
As \citet{gillespieGenerativeAIPolitics2024} writes in a similar study that explores social representation in LLM-generated narratives, ``Visibility matters...to the members of minorities and subcultures who long to see themselves represented... But visibility also matters for those outside of that marginalized community, even those who are indifferent to or averse to them. Greater visibility in media publicly acknowledges and legitimates non-normative identities and communities, helping to affirm that they too deserve a place in the social, political, and cultural landscape.''
The lack of explicit social representation in LLM outputs may rob marginalized communities of this important visibility and power, and it may rob readers of the opportunity to empathize with, and understand, those who are not like them.

To be clear, we believe the visibility of those who identify beyond a gender binary also greatly matters, especially at a time when reactionary laws, policies, and ideas threatening trans and non-binary people are spreading in the U.S. and in other parts of the world.
We applaud both the broader cultural shift toward using gender-neutral language, as well as the use of inclusive, respectful language that recognizes diverse gender expression.
However, we do not believe that the gender neutrality observed in these animal stories is a meaningful step toward gender diversity or inclusivity.
Rather we find that a concentrated focus on neutrality --- whether through ``it/its'' pronouns or the lack of gendered pronouns--- may not produce a genuinely balanced representational outcome, because feminine gender assignments are almost non-existent, and masculine assignments still persist. 
Hence, neutrality often functions in practice to amplify masculine bias, serving as a band-aid fix that fails to address underlying bias and more complicated alignment decisions.

Given our findings, we see an opportunity for more effective gender bias mitigation research in NLP. 
In particular, we suggest an alternative for gender finetuning that does not overly rely on gender-neutral language or gender avoidance. 
We encourage research that guides LLMs to produce a more equal distribution of genders (masculine, feminine, non-binary, trans, etc.) and ambiguity to better capture the nuances of language and identity.

\subsection{Limitations \& Future Work}
This study focuses on English-language stories and relies on grammatical structures that are specific to English and other languages that use referential gender.
Future work needs to be taken up to analyze how these dynamics play out in languages with grammatical gender, like Spanish, or languages that lack both grammatical and referential gender, like Estonian.
Additionally, 
in English, other words can convey gender identity beyond pronouns (e.g., wife, brother) \cite{caoGenderInclusiveCoreferenceResolution2021}, which may also be explored in future work.

Further, our study relies on prompting and specific prompt templates to evaluate the behavior of language models. While this is a common method for probing social understanding and biases in these systems, prompting can sometimes be unstable and has faced other criticism \cite{gao-kreiss-2025-measuring}.
We recognize this concern, but also note that most users now engage with LLMs via prompts, pointing to the usefulness of this approach. 

Another limitation of this work is our baselines. We do not directly compare LLM stories about animal characters with LLM stories about human characters. Additional paths for this work include a qualitative analysis on the theme, tropes, and plots within both the human-authored and LLM-generated narratives.
To strengthen our analysis, next steps could include conducting a similar survey with all $setting$ options.

Lastly, future work could also consider how these generated narratives impact readers; propose novel mitigation measures to better regulate this language; and even identify tropes and themes surrounding these animal characters, lending further insight into the ways neutrality bites.

\section{Conclusion}
In this paper, we use anthropomorphic fiction as a stress test for gender representation in contemporary LLMs. Across 23.8K story completions, we find that models frequently respond to gender ambiguity through \textit{avoidance}, either by repeating the animal noun or using neutral pronouns. This surface-level neutrality does not translate into equitable representation. When the models do explicitly assign gender, they overwhelmingly default to masculine protagonists, while feminine protagonists are nearly absent.
This experiment leads us to a broader claim about AI alignment with regard to gender and broader social representation:  
\textit{neutrality bites}.
In other words, we argue that prioritizing neutrality with regard to representations of gender, race, sexuality, or other identity categories, above all else, can function less as inclusion and more as erasure.

Our findings and broader argument have direct implications for the growing use of LLMs in story and fiction generation, including tools marketed for children like Google's Gemini Storybook, as well as tools marketed for adults and young adults, like Sudowrite or Character.AI.
Casting a character as gender neutral, or abstaining from gendering the character altogether, may seem like an effective way to reduce gender bias, but we show that, in practice, this strategy may result in imaginative worlds that have little to no feminine representation.
We show that if a parent or child generates an animal story with one of the world's leading AI models, they are much more likely to encounter a masculine character in a starring role rather than a feminine character, even though they will encounter many neutral or ambiguously gendered characters. 

We thus call for AI alignment approaches that move beyond gender neutrality, or representational neutrality, as the desired default in all ambiguous contexts. Concretely, we think model developers and auditors should (1) continue to evaluate bias in open-ended narrative settings where social categories must be inferred, as our experiment does; (2) measure representational parity in a way that includes, but does not prioritize, neutrality or ambiguity; (3) and explore post-training strategies that allocate social possibilities more equitably across fictional subjects rather than simply suppressing gendered or identity-related language.

Anthropomorphic stories offer a simple, reproducible audit setting for this work. 
We hope our study motivates more evaluations of how generative models imagine fictional worlds, human and otherwise, as they reflect and increasingly shape both the models and our own world.

\begin{acks}
We would like to thank \textit{The Pudding}, especially Russell Samora, for inspiring this project. We also thank the anonymous FAccT reviewers for their valuable insights and time. Lastly, we are grateful to the members of the University of Washington DSCO group for their feedback.
\end{acks}

\bibliographystyle{ACM-Reference-Format}
\bibliography{references}

\appendix
\section{Generative AI Usage Statement}
Generative AI was not used to write, edit, or revise this paper. Claude Code was used to assist in producing data visualizations and tables.

\section{FastCoref Performance}
To label our stories, we first used the FastCoref package \cite{otmazgin-etal-2022-f}. We manually annotated 200 randomly sampled stories and measured how our labels compared to the model's. On average, FastCoref achieved 0.95 accuracy.
\label{coref}
\begin{table}[h]
    \centering
    \begin{tabular}{l c c c}
    \hline
        Gender Category & Precision & Recall & F1 \\
        \hline
        Feminine & 0.80 & 1.00 & 0.89 \\
        Masculine & 1.00 & 1.00 & 1.00 \\
        Neutral & 1.00 & 0.89 & 0.94 \\
        Animal Name & 1.00 & 0.93 & 0.96 \\
        \hline
        \textbf{Weighted Avg.} & 0.95 & 0.95 & 0.95 \\
        \hline 
    \end{tabular}
    \caption{Precision, recall and F1 score from automatic gender labeling using FastCoref, using a random sample of 200 generated stories (50 per gender category).}
    \label{tab:accuracy}
\end{table}

\section{Survey Details}
\label{survey-dems}
The survey respondents from \citet{walsh_bearswillbeboys} were given the option to report their age and gender identity. Of the respondents:  578 identified as female, 413 as male, 46 as non-binary, and the remaining 290 chose not to disclose their gender. Additionally, 87 were under 20 years old, 347 in their 30s, 152 in their 40s, 66 in their 50s, and 68 above 60 years old; 261 respondents chose not to disclose their age.

\section{Results Continued}
\label{appendix}
\begin{figure}[!h]
    \centering
    \includegraphics[width=0.9\linewidth]{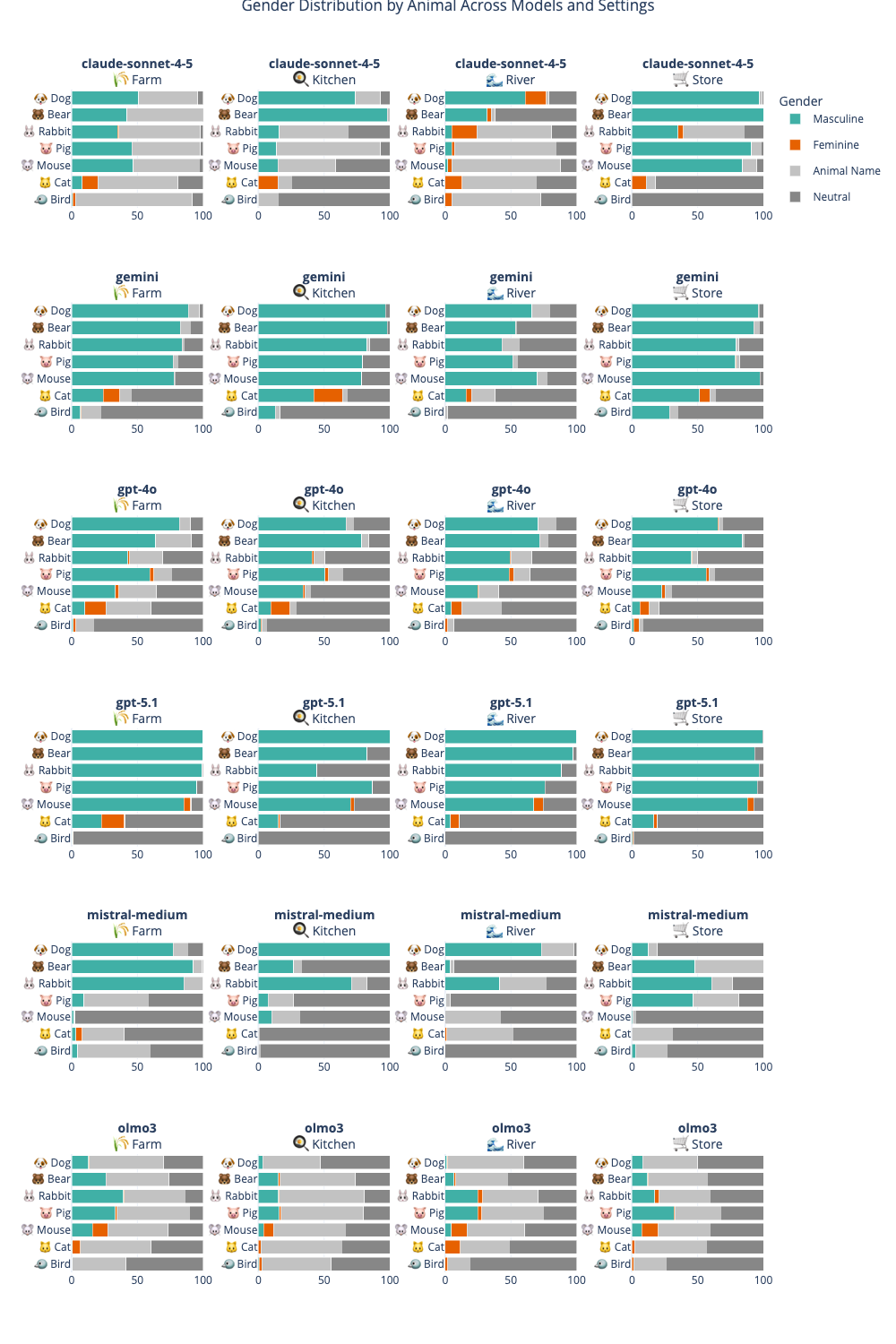}
    \caption{Gender assignment distribution across narrative settings for generated stories.}
    \label{fig:settings}
\end{figure}

\begin{figure}[!h]
    \centering
    \includegraphics[width=0.8\linewidth]{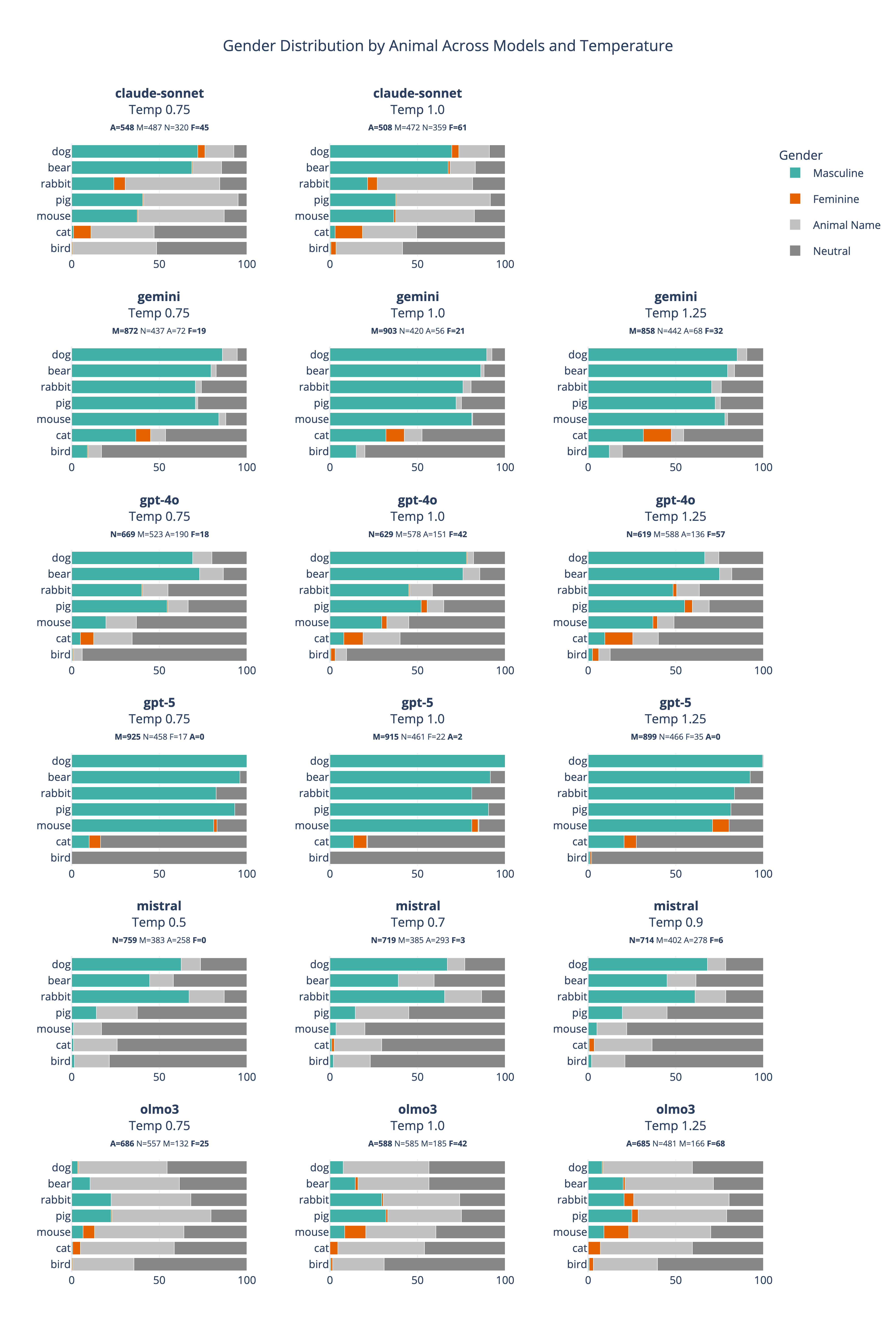}
    \caption{Gender assignment distribution across temperatures for generated stories. }
    \label{fig:temps}
\end{figure}

We compare model performance of a story-completion task across narrative setting and temperature in Figure \ref{fig:settings}
and \ref{fig:temps}.

\end{document}